\documentclass[letterpaper, 10 pt, conference]{ieeeconf}  

\IEEEoverridecommandlockouts                              

\usepackage{amsmath} 
\usepackage{amssymb}  
\usepackage{graphicx}
\usepackage{multirow}

\newcommand{\etal}{\textit{et al}.}
\newcommand{\ie}{\textit{i}.\textit{e}. }
\newcommand{\eg}{\textit{e}.\textit{g}. }
\newcommand{\cf}{\textit{c}.\textit{f}. }
\newcommand{\etc}{\textit{etc}. }

\title{\LARGE \bf
Hallucinating Beyond Observation: Learning to Complete \\ with Partial Observation and Unpaired Prior Knowledge
}

\author{Chenyang Lu and Gijs Dubbelman
	\thanks{The authors are with the Mobile Perception Systems research cluster of the SPS/VCA group, Dept. of Electrical Engineering,  Eindhoven University of Technology, The Netherlands.
		{\tt \{c.lu.2, g.dubbelman\}@tue.nl}}%
}

\begin{document}

\maketitle
\thispagestyle{empty}
\pagestyle{empty}

\begin{abstract}
	
	We propose a novel single-step training strategy that allows convolutional encoder-decoder networks that use skip connections, to complete partially observed data by means of hallucination. This strategy is demonstrated for the task of completing 2-D road layouts as well as 3-D vehicle shapes. As input, it takes data from a partially observed domain, for which no ground truth is available, and data from an unpaired prior knowledge domain and trains the network in an end-to-end manner. Our single-step training strategy is compared against two state-of-the-art baselines, one using a two-step auto-encoder training strategy and one using an adversarial strategy. Our novel strategy achieves an improvement up to +12.2\% F-measure on the Cityscapes dataset. The learned network intrinsically generalizes better than the baselines on unseen datasets, which is demonstrated by an improvement up to +23.8\% F-measure on the unseen KITTI dataset. Moreover, our approach outperforms the baselines using the same backbone network on the 3-D shape completion benchmark by a margin of 0.006 Hamming distance.
	
\end{abstract}

\section{Introduction}
\label{sect_intro}

The ability to understand the scene beyond what is directly observed, is a fundamental cognitive ability. Conceptually, it requires projecting learned knowledge onto current ego-centric observations, which inherently have a limited field-of-view (FOV) and contain occlusions, to complete the scene. Such scene completion is an essential task for applications of computer vision that are concerned with, for example, mobile robotics and intelligent self-driving vehicles.

\begin{figure}[!tbp]
	\centering
	\includegraphics[width=\linewidth]{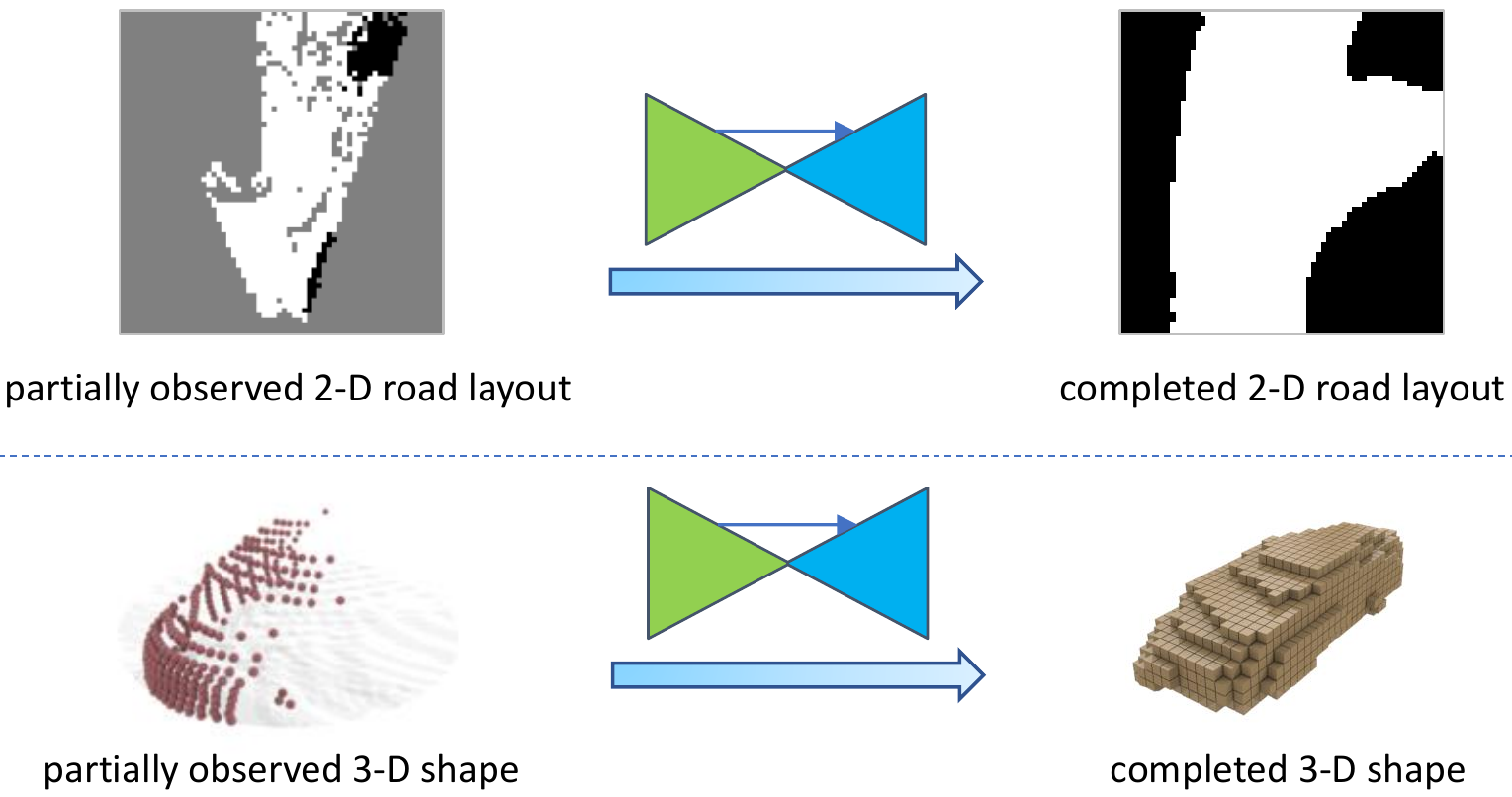}
	\caption{System overview. The convolutional encoder-decoder network with skip connections takes the observed incomplete road layout or vehicle shape as input and hallucinates a completed one.}
	\label{fig_system_example}
\end{figure}

Towards this goal of scene completion beyond the ego-centric observations, we propose a novel training strategy that allows convolutional encoder-decoder networks with skip connections to hallucinate and complete: 1) 2-D road layouts, and 2) 3-D vehicle shapes \cite{Stutz2018}, from partial observations, see Figure \ref{fig_system_example}. Our training strategy takes as input data from a partially observed domain, for which no ground truth is available, and data from a prior knowledge domain, which has no direct one-to-one correspondence with the partially observed domain. We call this \textit{unpaired} prior knowledge and we show that our novel end-to-end training strategy allows the network to outperform the state-of-the-art \cite{Schulter2018, Stutz2018} in several metrics.

The connection with related tasks, such as image inpainting, scene hallucinating, and road layout understanding is addressed in Section \ref{sect_ralate_work}, and the representation of our novel 2-D road layout hallucinating, including partially observed and prior knowledge data, is discussed in Section \ref{sect_road_layout_representation}. One state-of-the-art \cite{Stutz2018}, which hallucinates from partially observed data using unpaired prior knowledge, is a variational auto-encoder (VAE) \cite{Kingma2013} trained in two steps: 1) first the encoder and decoder are trained on the unpaired prior knowledge, 2) then the encoder is re-trained on the partially observed data while keeping the decoder fixed. A fundamental limitation of this approach, which we use as the first baseline for our experiments, is its inability to leverage skip connections between the encoder and decoder, which could greatly improve the level of detail in the decoded output. As the second baseline, we use a state-of-the-art method that can utilize skip connections by applying a generative adversarial network (GAN) training strategy \cite{Schulter2018}, which has the typical challenges of GAN training instability \cite{Lucic2017}.

\begin{figure*}
	\begin{center}
		\includegraphics[width=\linewidth]{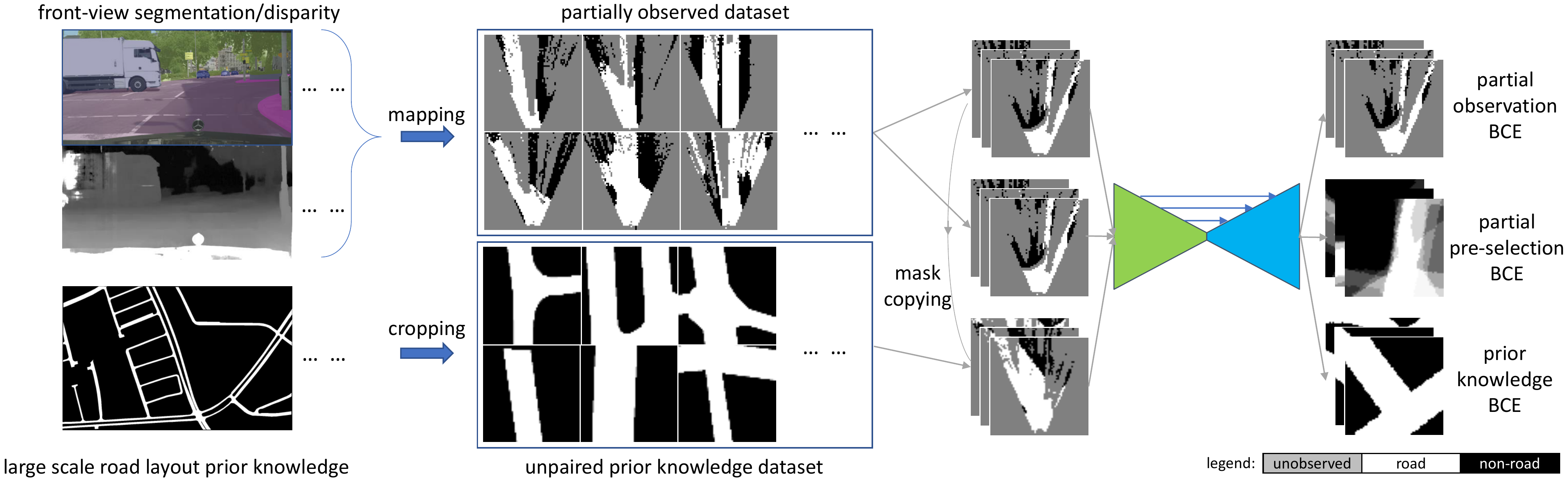}
	\end{center}
	\caption{The proposed approach taking 2-D road layout hallucinating as an example. Left: the generation of the partially observed and the unpaired prior knowledge road layout datasets. Right: our single-step training approach. The real (un)observation masks are copied onto the unpaired prior knowledge samples as one of the inputs, and three supervisions are applied simultaneously during training. See Section \ref{subsect_training_with_multi} for details. Legend: BCE = binary cross entropy.}
	\label{training}
\end{figure*}

Our proposed single-step training strategy, see Figure \ref{training} and Section 4, does not suffer from the aforementioned limitations. Therefore, it can unlock the benefits of fully convolutional networks that employ skip connections, without the need for adversarial supervisions. These benefits, which are demonstrated by multiple experiments in Section 5, include: 1) a higher level of detail in the decoded output, and 2) a higher level of generalization to unseen data.

To summarize, we make the following contributions: 
\begin{itemize}
	\item A one-step training strategy for the hallucinating task that simultaneously utilizes knowledge from a partially observed domain without ground truth and an unpaired prior knowledge domain.
	\item In this setting, we are the first to benchmark using two independent hallucinating tasks, i.e. 2-D road layout hallucinating and 3-D vehicle shape completion. 
	\item Using these benchmarks, we demonstrate that our proposed one-step training strategy outperforms current state-of-the-art \cite{Schulter2018, Stutz2018} in terms of level-of-detail and generalization for unseen data.
\end{itemize}

For 2-D road layout hallucinating, we develop a novel benchmark that is shared with the research community \cite{Lu2019iccv}. As this benchmark poses other challenges than the existing 3-D vehicle shape completion benchmark of \cite{Stutz2018}, we primarily use this novel benchmark to explain and illustrate our one-step training strategy in this work.

\section{Related work}
\label{sect_ralate_work}

\textbf{Image inpainting: } Closely related to our task of hallucinating the road layout is image inpainting. In recent years, deep convolutional neural networks (CNNs) enable the possibility of  image inpainting with large missing areas, as CNNs can extract abstract semantic information from the observable context. The Context Encoder (CE) \cite{Pathak2016} network is proposed to inpaint the image with large rectangle areas missing at the image center by applying reconstruction and adversarial loss \cite{Goodfellow2014a} in training. CE-like networks \cite{Demir2018, Iizuka2017, Li2017d} are proposed with additional discriminative networks applied on locally missing regions, or on the entire image in a patch-wise manner, which are able to perform inpainting with regions missing at arbitrary position. Given a trained generative network, Yeh \etal \cite{Yeh2016} do inpainting by finding the embedding vector that minimizes the reconstruction loss by applying back-propagation to the input embedding vector. The main difference between inpainting and the tasks addressed by us and \cite{Schulter2018, Stutz2018}, is that the previously introduced methods of image inpainting are all in a setting in which the complete ground truth is available.

\textbf{Scene hallucinating: } If the ground truth is not available, one has to hallucinate the region to be completed. Srikantha and Gall \cite{Srikantha2016} propose a system to hallucinate a depth map and a semantic map, given an RGB image and a noisy, incomplete depth map, which is able to remove the foreground objects. Schulter \etal \cite{Schulter2018} proposes a CNN to conduct a similar task without depth, by intentionally adding random foreground masks during training. Recently, a VAE with two-step training \cite{Stutz2018} is proposed for shape completion. In the first step, a canonical VAE is trained on a complete shape prior dataset which has no direct correspondence to the incomplete shape dataset. Then the amortized maximum likelihood (AML) is applied as supervision on the incomplete shape data with the decoder fixed in the second training step. This VAE approach is originally used to learn 3-D vehicle shape completion, but it can be generalized to similar tasks such as our 2-D road layout hallucinating. Therefore, we use this approach as a baseline, which is referred to as \textit{AML baseline}.

\textbf{Road layout understanding: } Road layout hallucinating can be seen as a specific approach of road layout understanding, which is an important task for robot and intelligent vehicle navigation. One challenge is that the ego-centric sensory data usually contains occlusions of the foreground objects, which makes roads visually incomplete. Many works tackling occlusions are focusing on using front-view images, such as road boundary detection \cite{Amayo2018}, and road segmentation \cite{Becattini2018}. Less work has been carried out on the top-view ego-centric sensing. In \cite{Schulter2018}, the proposed system can produce a top-view road layout, while the occlusion is still addressed on the front-view image by pre-processing. In the later top-view refinement, the GPS is heavily relied on for a paired reconstruction supervision. We use a variant of this method, which only uses unpaired prior knowledge and thus no GPS pairing, as the second baseline. It is referred to as \textit{GAN baseline}.

\section{Road layout data representation}
\label{sect_road_layout_representation}

As we use 2-D road layout hallucinating as the primary task to illustrate our novel one-step training strategy in Section \ref{sec_learning}, we first introduce some required data representations of our novel benchmark. See the left part of Figure \ref{training}. 

\subsection{Partially observed dataset}

Our road layout representation is derived from \cite{Lu2018a}, where the environment in front of the vehicle is divided into 2-D semantic grids in Cartesian coordinates. As the map is generated based on one-shot sensing, some grids are always out of the camera's FOV and some are missing due to the occlusion of the foreground objects and noises. In this work, the value of each grid indicates the \textit{status} of the road: 1, 0, 0.5 represent \textit{road}, \textit{non-road}, and \textit{unobserved}, respectively. Given a front-view image from the camera mounted on the vehicle, its corresponding semantic map, the camera calibration data, and the depth/disparity map, a 3-D point cloud with semantics is generated. The points are mapped to the 2-D ground plane and the grid cells are filled with their corresponding \textit{status}, based on the semantic label statistics of the points contained in the grid cell (majority vote). Specifically, the grid is assigned with 1 if the majority label is \textit{road}, 0 if the majority label is related to the static world except road (\eg construction, sidewalk, \etc), and 0.5 if the majority label belongs to movable foreground objects (\eg vehicle, human, \etc) or if the majority label does not exist, due to the camera's FOV limitation or occlusion. 


\begin{figure}[!tbp]
	\begin{center}
		\includegraphics[width=\linewidth]{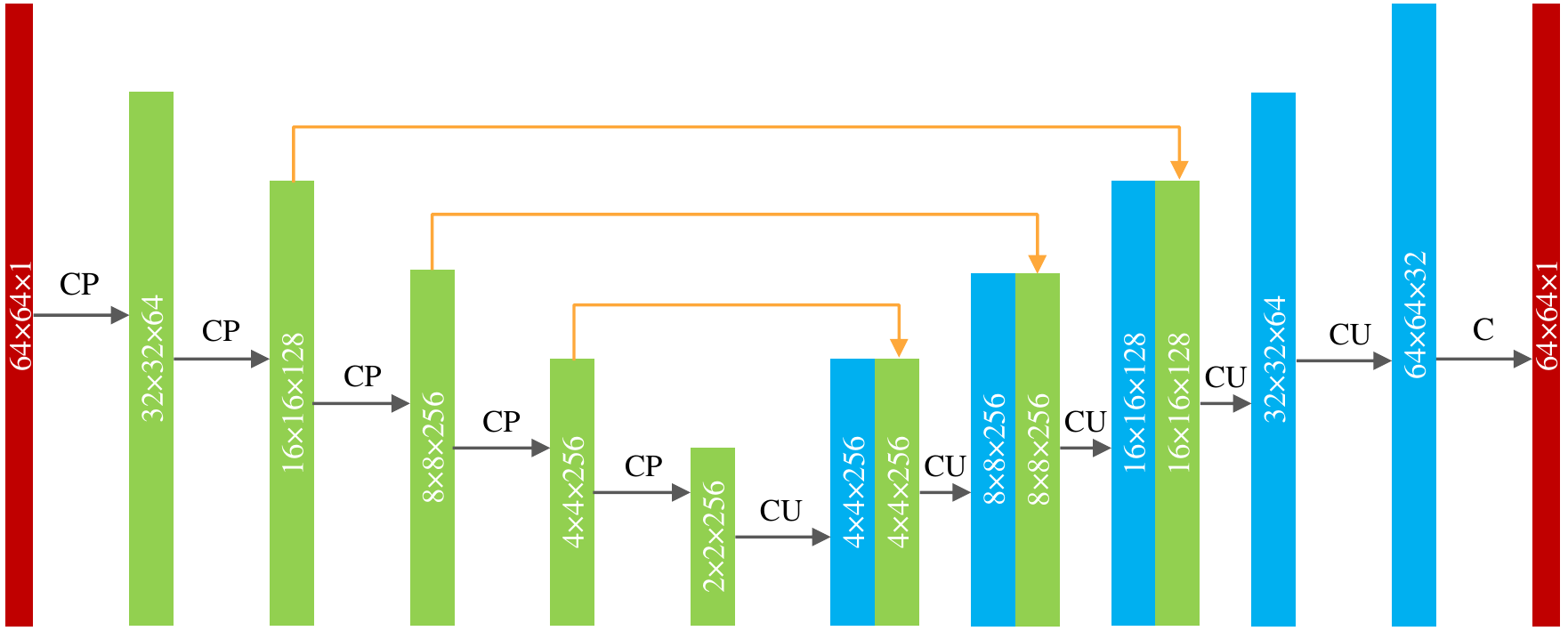}
	\end{center}
	\caption{The network for 2-D road layout data. Each black arrow represents a network module, and each block represents a tensor. Legend: CP = two U-net-like \cite{Ronneberger2015} convolutional layers (with batch normalization) following a 2*2 max pooling layer, CU = two U-net-like \cite{Ronneberger2015} convolutional layers (with batch normalization) following an up-sample layer, C = the last convolutional layer with kernel size 3 that generates the final output with a sigmoid layer.}
	\label{fig_net}
\end{figure}

\subsection{Unpaired prior knowledge dataset}

In contrast to the partially observed dataset, which contains the real semantic-metric road information, the \textit{unpaired prior knowledge dataset} contains no real observation but a binary prior knowledge of what a road should normally look like in a top-down view. The underlying assumption is that the road layout representation follows a certain distribution in a latent low dimensional space. In the case that the partially observed road layout can provide enough information, one can predict the overall road layout, which aligns with the partial observation and that also follows the distribution of the prior knowledge. Given a binary road segmentation in a top-down view of a region, we use a sliding window to crop the large road map into smaller ones with the same size as the samples in the partially observed dataset, see the left part of Figure \ref{training} for an example. There exists no one-to-one correspondence between samples in the prior knowledge dataset and the partially observed dataset.

\begin{figure*}
	\begin{center}
		\includegraphics[width=\linewidth]{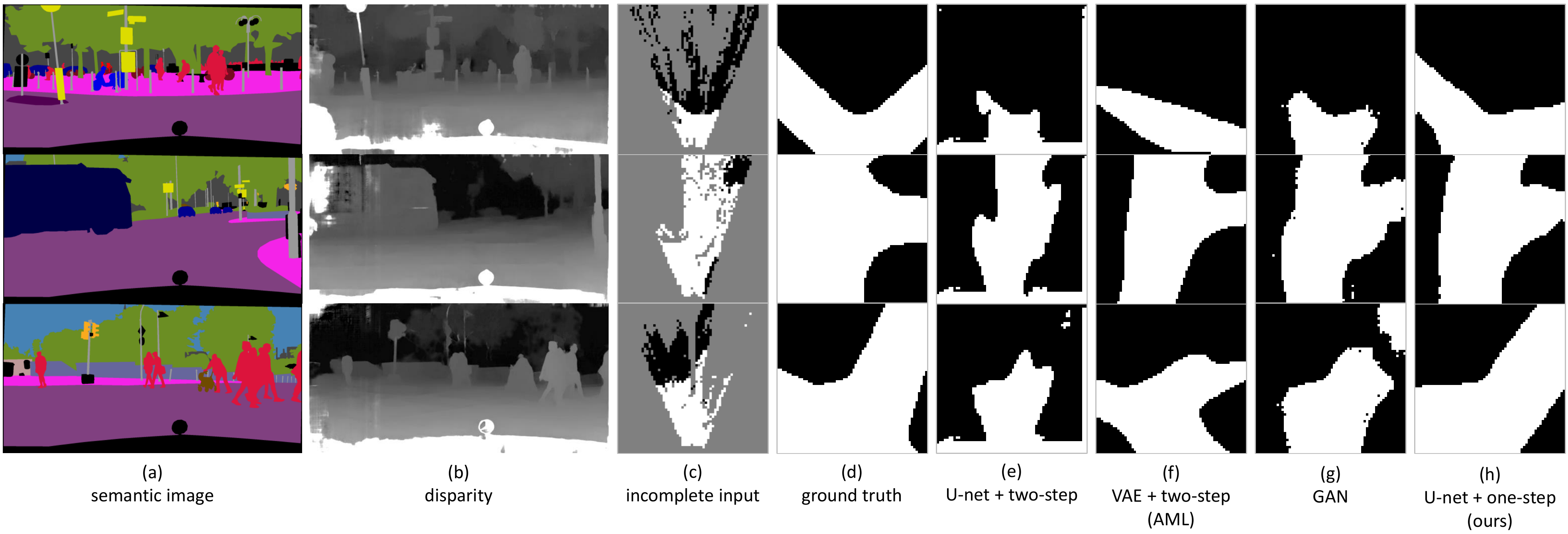}
	\end{center}
	\caption{The inputs and outputs of the 2-D road layout hallucinating task. (a) and (b) are the semantic map and the disparity map used for partially observed map generation. (c) is the generated partially observed input road layout. (d) is the manually annotated ground truth based on the visual cue of the front-view image and the corresponding satellite image. (e)-(h) are the outputs of the U-net-like network with two-step training, AML baseline \cite{Stutz2018}, GAN baseline \cite{Schulter2018}, and the U-net-like network with one-step training (ours), respectively.}
	\label{fig_data_example}
\end{figure*}

\section{Learning to complete}
\label{sec_learning}

Our proposed convolutional encoder-decoder network learns to complete the incomplete input data by training jointly on partially observed and unpaired prior knowledge datasets. We use multiple supervision signals for different datasets and train our network in a single step. In this section, we present our approach using the road layout hallucinating task as the working example.

\subsection{Network structure}
\label{subsec_network_struct}

A U-net-like \cite{Ronneberger2015} fully convolutional encoder-decoder network with \textit{skip connections} is utilized for the hallucinating task, see Figure \ref{fig_net} for the network that processes the road layout data. Given the incomplete data $M$ as input, the encoder produces a high-level feature map at the bottleneck, as well as the intermediate feature maps from the last three max pooling layers before the bottleneck, which are fed into the decoder by means of the skip connections. In the decoder, the feature map is convolved and upsampled with nearest-neighbor interpolation, which leads to the final completed sigmoid prediction $\hat{M}$. The intermediate feature maps from the encoder are concatenated with the decoder's intermediate feature maps at the corresponding layers, which enables the direct transfer of the accurate observed detail information to the final completed outputs.

Note that the proposed U-net-like structure cannot be trained in AML baseline's two-step manner, as the decoder will only learn a simple copy-paste task on the prior knowledge set due to the skip connections, which is demonstrated with an experiment in Section 5. Unlike the AML baseline, the second adversarial training based \textit{GAN baseline} can work with the usage of the skip connections, but the performance improvement is limited, as shown in the experiments. To unlock the power of skip connections, we propose the following novel supervisions for a better hallucinating performance.

\begin{table*}
	\renewcommand{\arraystretch}{1.2}
	\caption{Quantified performance of three methods evaluated on both 2-D road layout hallucinating (using Cityscapes dataset \cite{Cordts2016a}) and 3-D vehicle shape completion (using SN-clean dataset \cite{Stutz2018}) tasks. See Figure \ref{fig_data_example} and \ref{fig_shape_completion} for their corresponding qualitative results. The method of U-net-like network with two-step training is not evaluated as it cannot perform the hallucinate task: see  Figure \ref{fig_data_example}(e), the region which is out of FOV and tends to copy-paste the observed road. *Two conditions are reported in \cite{Stutz2018}, please see \cite{Stutz2018} for the details.}
	\label{tab_main_compare}
	\begin{center}
		\begin{tabular}{c||ccc|cc|cc||c}
			\hline
			\multirow{3}{*}{\bfseries method} & \multicolumn{7}{c||}{\bfseries 2-D road layout}  & \bfseries 3-D vehicle shape\\
			\cline{2-9}
			& \multicolumn{3}{c|}{\bfseries contour (no relaxation)}	&\multicolumn{2}{c|}{\bfseries full pixels} &\multicolumn{2}{c||}{\bfseries undetected pixels} & \bfseries Hamming distance\\
			\cline{2-8}
			& \bfseries precision& \bfseries recall& \bfseries $F$-measure& \bfseries pixel acc.  & \bfseries mean IoU& \bfseries pixel acc. & \bfseries mean IoU & \bfseries (lower is better) \\
			\hline\hline
			\bfseries AML& 20.4 & 19.9 & 20.0 & 86.7 & 75.8 & 82.2 & 65.8 & 0.041 (0.043)*\\
			\bfseries GAN & 23.9 & 31.1 & 26.6 & 87.6 & 76.9 & 82.7 & 65.7  & 0.043 \\
			\bfseries ours & \bfseries 32.5 & \bfseries 32.5 & \bfseries 32.2 & \bfseries 88.4 &  \bfseries 78.5 & \bfseries 83.7 & \bfseries 68.6 & \bfseries 0.035\\
			\hline
		\end{tabular}
	\end{center}
\end{table*}

\begin{figure}
	\begin{center}
		\includegraphics[width=\linewidth]{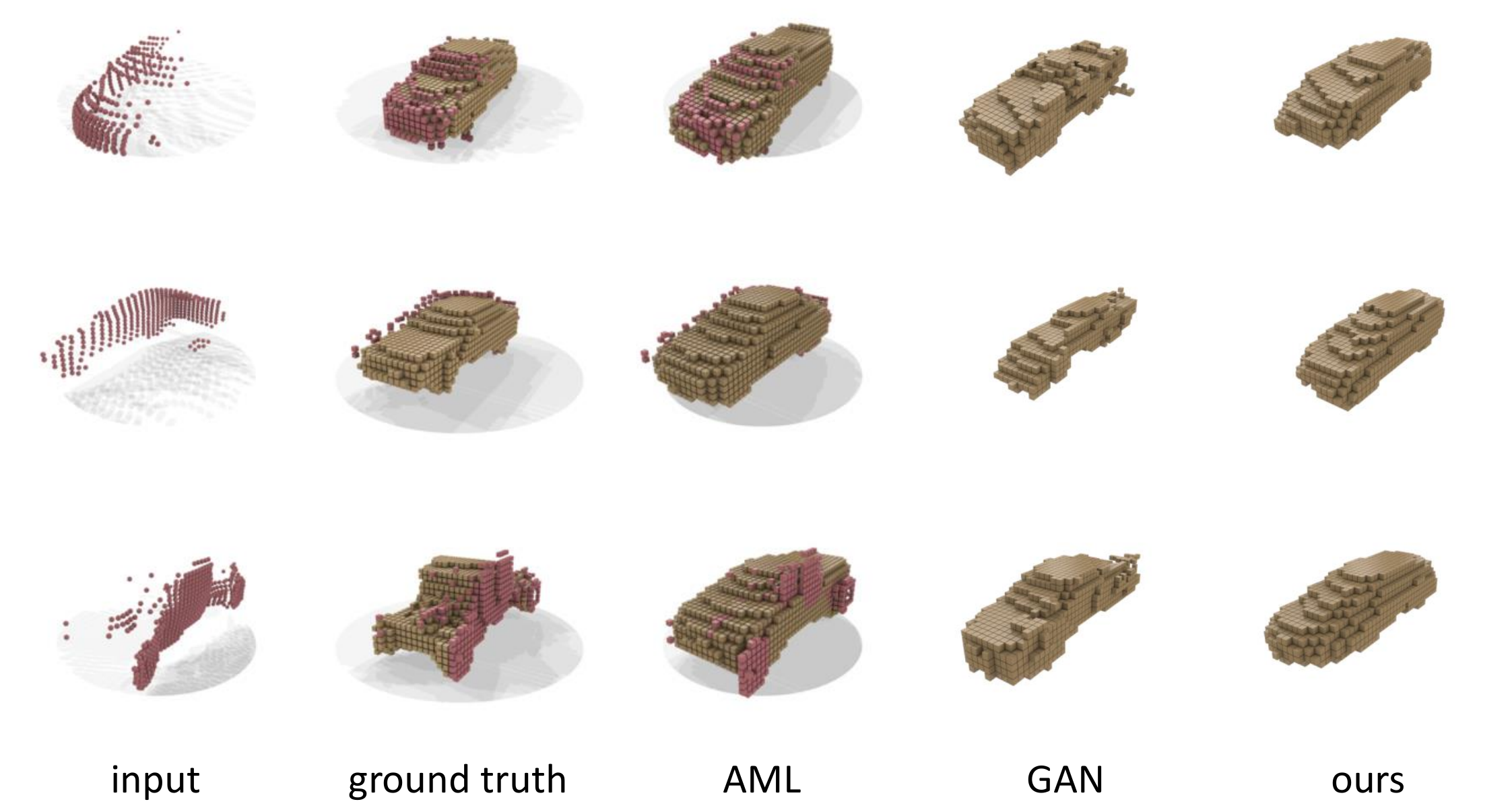}
	\end{center}
	\caption{The inputs and outputs of the 3-D vehicle shape completion task. The input, ground truth and AML baseline samples are from \cite{Stutz2018}.}
	\label{fig_shape_completion}
\end{figure}

\subsection{Training with multiple input-target pairs} 
\label{subsect_training_with_multi}

We propose to train our U-net-like model in one step, with three input-target pairs applied for each iteration. 

\textbf{Observation pair:} The first input-target pair, called \textit{observation pair}, is from the partially observed dataset. The goal is to encourage the network to reconstruct the observed regions, and the pixels that are not observed are ignored in training. Clearly, this pair will not influence the network's behavior at the unobserved region, which will be addressed by the next two input-target pairs. 

\textbf{Pre-selection pair:} In the second input-target pair, the input is retained as in the first one, while the target is selected from the prior knowledge dataset. The underlying assumption is that if there exists a prior knowledge dataset with infinite number of samples, one can find one sample that aligns with the unknown ground truth. In the real case, the prior knowledge dataset is limited and in general does not contain samples that exactly match the unknown ground truth for the partially observed dataset. Therefore, we select a certain number of samples from the unpaired prior knowledge dataset that has the highest mean intersection-over-union (IoU) evaluated on the observed pixels. The selected samples are averaged and used as the target of the second input-target pair, which is named \textit{pre-selection pair}. It provides explicit supervision on both observed and unobserved regions. 

\textbf{Masked prior knowledge pair:} In the third pair, the unobserved masks in the partially observed dataset are copied and applied onto the prior knowledge samples. In this case, the input is intentionally masked with the realistic (un)observation patterns, and the target is the fully observed prior knowledge sample. This encourages the network to learn the pattern association between the unobserved mask, and the underlying ground truth behind the mask by automatically generating partially observed and complete training pairs from prior knowledge dataset, which is referred as \textit{masked prior knowledge pair}. 

With these different input-target training pairs, the connection between the two datasets is constructed and the knowledge is transferred and shared in one-step training.

\subsection{Multiple supervisions} 
When training our U-net-like network, three losses for the three input-target pairs are applied in training, which are based on the binary cross entropy (BCE). For each pixel prediction $\hat{m}$ and its corresponding target $\bar{m}$, the BCE can be computed. 
We denote the predicted road layout in three information flows as $\hat{M}_1$, $\hat{M}_2$, and $\hat{M}_3$, respectively, and the target layout is denoted as $\bar{M}_1$, $\bar{M}_2$, and $\bar{M}_3$. The losses are in the same form:
\begin{equation}
Loss_{i} (\hat{M}_i, \bar{M}_i) = \frac{1}{N_{i}}\sum_{\bar{m} \in \text{valid}} BCE(\hat{m}, \bar{m})
\end{equation}
where $i=1, 2, 3$ and $N_{i}$ is the number of pixels which are \textit{valid}: in the observation pair's loss, the observed pixels are valid; in the pre-selection pair's loss, only the pixels that all pre-selected samples agree on (the average is strict 0 or 1) are valid; and in the masked prior knowledge pair's loss, all pixels are valid. We denote them as \textit{partial observation BCE}, \textit{partial pre-selection BCE}, and \textit{prior knowledge BCE} in the figures. The overall loss can be expressed as
\begin{equation}
Loss = \lambda_1 \cdot Loss_1 + \lambda_2 \cdot Loss_2 + \lambda_3 \cdot Loss_3
\end{equation} 
where $\lambda_1$, $\lambda_2$, $\lambda_3$ are the weights for balancing the losses.

\begin{figure}[!tbp]
	\begin{center}
		\includegraphics[width=\linewidth]{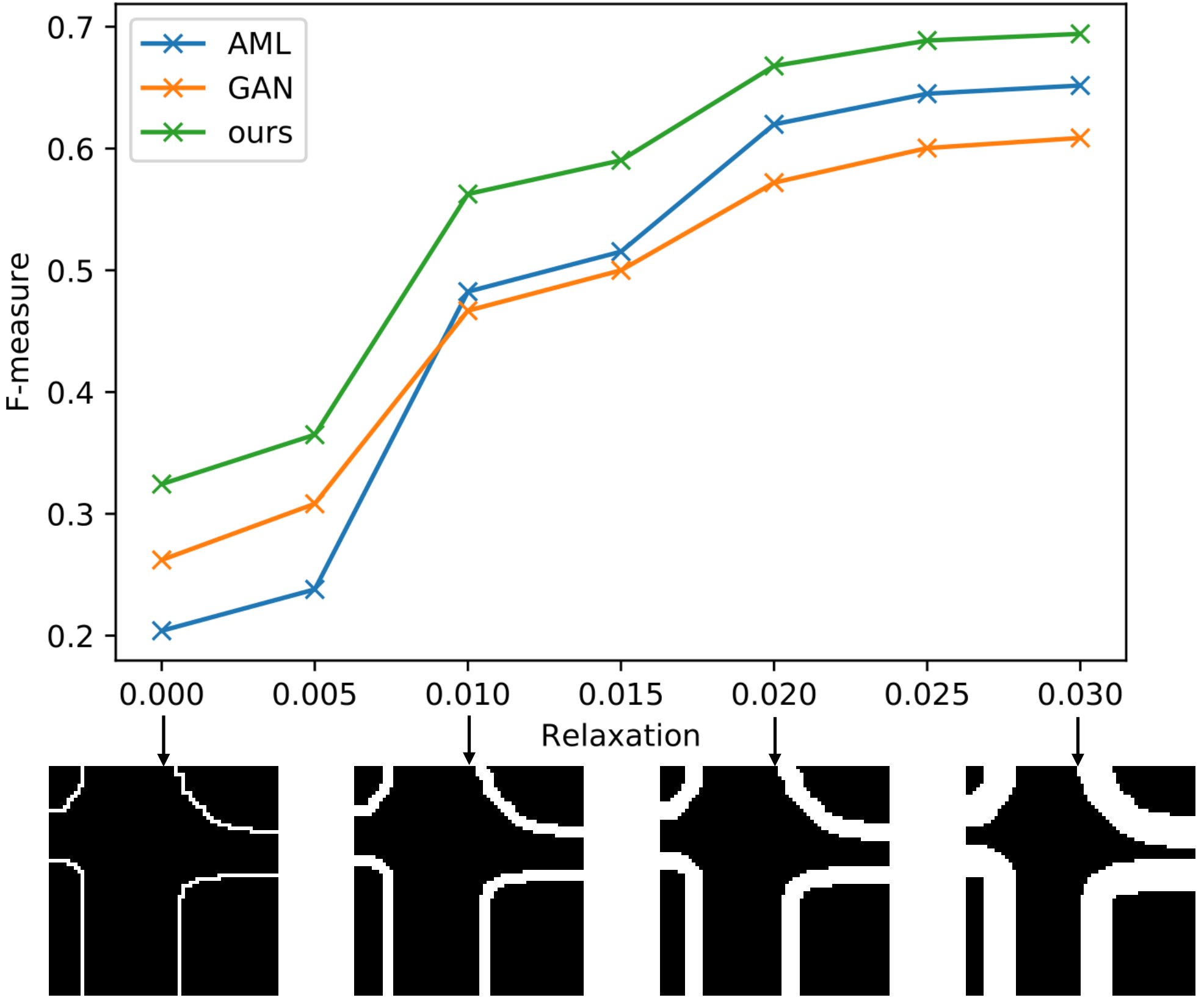}
	\end{center}
	\caption{On the 2-D road layout hallucinating task, we plot $F$-measure in different boundary relaxation settings, where our approach is consistently having higher $F$-measures. The x-axis represents the degree of relaxation (higher indicates more tolerance of the road boundary matching). As visualized below, the accepted region of the boundary is enlarged when the relaxation parameter increases.}
	\label{fig_relaxation_compare}
\end{figure}

\section{Experiments}

We conduct the following three experiments to evaluate the performance of different approaches:
\begin{itemize}
	\item \textbf{Quantitative evaluation:} We evaluate the quantitative performance of the baselines and the proposed approach on both 2-D road layout hallucinating and 3-D vehicle shape completion tasks;
	\item \textbf{Ablation studies:} We compare the performance with certain input-target pairs ignored to verify the effectiveness of the proposed training strategy on both tasks.
	\item \textbf{Generalizability evaluation:} On the 2-D road layout hallucinating task, we evaluate the performance of the baseline and our approach on the unseen KITTI \cite{Geiger2013a} and unseen hold-out masked prior knowledge dataset.
\end{itemize}

\subsection{Experimental settings}
\label{subsect_experimental_settings}

\textbf{Datasets:} For the \textit{2-D road layout hallucinating} task, we use two publicly available datasets for the generation of partially observed data, namely Cityscapes \cite{Cordts2016a} and KITTI \cite{Geiger2013a}. Cityscapes dataset comes with 2975 training samples and 500 validation samples with publicly available fine semantic annotations, the corresponding disparity maps and calibration data. We use the validation set for tests in our experiments. The KITTI semantic dataset contains 200 available annotated training samples. Considering the limited amount of samples, we only use it for additional experiments of generalizability. As for prior knowledge dataset generation, we use the AerialKITTI dataset \cite{Mattyus2015}, which contains 20 large annotated top-view road layout binary images. About 154K prior knowledge road layout samples are generated. Unlike other image inpainting tasks, the ground truth, i.e. the actual road layout, for the front-view Cityscapes and KITTI images is not available. However, to be able to evaluate the performance, 162 samples from Cityscapes and 142 samples from KITTI are manually annotated as the ground truth by analyzing the visual cues from front-view images and the corresponding satellite images given the GPS signals. These annotations are shared with the community \cite{Lu2019iccv}.

For the \textit{3-D vehicle shape completion} task, the used dataset \cite{Stutz2018} contains two parts: the reference shapes (prior knowledge), and the ground truth shapes, which are utilized to generate partially observed samples, with (SN-noisy) and without (SN-clean) noises. Being different from our dataset, in this case, the prior knowledge and partial observation samples are from an identical distribution, \ie ShapeNet \cite{Chang2015}, which is not the case for our novel 2-D road layout hallucinating benchmark. Please see \cite{Stutz2018} for more details of this dataset.

\textbf{Baselines:} For both tasks, we consider two baselines in our experiments, namely AML \cite{Stutz2018} and GAN \cite{Schulter2018}, as introduced in Section \ref{sect_intro} and \ref{sect_ralate_work}. For a fair comparison, we keep the network settings of the baselines and the proposed network as similar as possible. The necessary differences exist between the AML baseline and our model, including 1) the fully connected layer at the baseline's bottleneck is deleted in our network, and 2) in the decoder, the input channels of three convolutional and upsampling modules are increased for the usage of skip connections, while the output channels of the first convolutional layer in these modules remain the same. An identical model is used for the GAN baseline and our approach, for a valid comparison between different supervisions. For the discriminator of the GAN baseline and its training strategy, we follow the settings of CycleGAN \cite{Zhu2017}, as they are conducting a similar translation task. Please see our supplementary material for more details.

\textbf{Training details:} We train the networks for two tasks in a similar manner using Pytorch \cite{Paszke2017}, and here we take the 2-D road layout hallucinating task as an example. To train the AML baseline model in our setting, in the first step, the BCE and Kullback-Leibler (KL) divergence are used for training on the prior knowledge samples for 20 epochs. Afterwards, the partial observation BCE and KL divergence are used for re-training the encoder on the Cityscapes partially observed dataset for 60 epochs. The size of the embedding vector is 256, and we observe similar results with either larger or smaller size. The GAN baseline is trained for 60 epochs in one step, and the prior knowledge samples are used for discriminator learning. The proposed approach is trained with 60 epochs by feeding the same prior knowledge and the same partially observed datasets together. For the sake of efficiency, we randomly choose a subset of the prior knowledge training set (50K) for targets pre-selection. The weights of three losses $\lambda_1$, $\lambda_2$, $\lambda_3$ are set as 0.5, 0.25 and 0.25, and we observe that our model is insensitive to the small weights' variance. We use Adam \cite{kingma2014} optimizer for training all the models. As the partially observed dataset contains noise from imperfect disparity maps, we randomly inverse the binary pixel values on the observed pixels of the input maps with 15\% probability, as a data augmentation in both models. More training details of two tasks can be found in our supplementary material.

\begin{table*}
	\renewcommand{\arraystretch}{1.2}
	\caption{Quantified performance of the proposed approach and that with certain supervision signals ignored. As some performance differences are with small margins, for the metrics of 2-D road layout data, we train the network 5 times with different random seeds and compute the average for each metric.}
	\label{tab_ablation}
	\begin{center}
		\begin{tabular}{l||c|c|c||c}
			\hline
			\multirow{3}{*}{\bfseries method} & \multicolumn{3}{c||}{\bfseries 2-D road layout} & \bfseries 3-D vehicle shape\\
			\cline{2-5}
			& \multicolumn{1}{c|}{\bfseries contour }	&\multicolumn{1}{c|}{\bfseries full pixels} &\multicolumn{1}{c||}{\bfseries undetected pixels} & \bfseries Hamming distance\\
			&  \bfseries $F$-measure & \bfseries mean IoU & \bfseries mean IoU & \bfseries (lower is better) \\
			\hline
			\hline
			\bfseries ours & \bfseries 32.2& 78.5 & 68.6 & \bfseries 0.035\\
			\bfseries ours w/o observation pair & 29.8 & \bfseries 78.8 & \bfseries 69.0 & 0.042\\
			\bfseries ours w/o pre-selection pair& 31.6 & 78.0 & 68.2 &0.039\\
			\bfseries ours w/o masked prior knowledge pair & 31.5 & 78.5  & 67.6& \bfseries 0.035\\
			\bfseries only pre-selection pair & 24.0 & 77.9  & 67.4 & 0.040\\
			\bfseries only masked prior knowledge pair & 30.1 & 78.7  & \bfseries 69.0 & 0.043\\
			\hline
		\end{tabular}
	\end{center}

\end{table*}

\subsection{Quantitative evaluation}

We analyze the results of the baselines and our proposed approach on two different tasks, and show that our approach consistently outperforms the other baselines.

\textbf{2-D road layout hallucinating:} Two aspects of performance are evaluated: segmentation accuracy and contour accuracy. The segmentation accuracy is evaluated on all pixels as well as on the unobserved pixels. As the output of our task is a binary mask indicating the road and non-road, we use pixel accuracy and mean IoU averaged over samples for comparison. Furthermore, we use contour accuracy to evaluate the quality of road boundary prediction. This metric is derived from the metrics in DAVIS segmentation challenge \cite{Perazzi2016}. The idea is to focus on the performance of whether the boundary is successfully predicted in terms of precision $P$, recall $R$, and their corresponding $F$-measure with $F = \frac{2 \cdot P \cdot R}{P + R}$. This metric is more important for our task because the accuracy of the road boundary is essential for the navigation in autonomous driving. In our experiments, the boundary region matching can be relaxed by morphological operations, and the performances in different relaxation conditions are reported. 

We observe that the U-net-like network with two-step training cannot conduct the hallucinating task, as explained in Section 4.1, \cf Figure \ref{fig_data_example}(f), and is not further quantitatively evaluated. Table \ref{tab_main_compare} presents the performance of two baselines and our approach evaluated on Cityscapes. The GAN baseline exhibits slightly better results than the AML baseline, while our approach outperforms both two baselines by more than 0.8\% in pixel accuracy and 2.8\% in mean IoU. More importantly, our approach performs better than two baselines in term of contour accuracy with large margins: nearly 12\% and 9\% in $F$-measure, and the performance gap is consistent with boundary relaxation, see Figure \ref{fig_relaxation_compare}. With the relative shallow features directly fed into the decoder by the skip connections, it is easier for the network to predict the road layout whose boundary is better aligned with the corresponding partially observed inputs. In addition, with the identical network with skip connections, our proposed training strategy boosts the contour accuracy more significantly than the GAN based baseline, which performs even worse than the AML baseline (without skip connections) under large relaxation conditions, \cf Figure \ref{fig_relaxation_compare}. 

Interestingly, we notice that combining three proposed supervisions and adversarial supervision leads to performance degradation. We also observed that random seeds play an important role when training the GAN baseline: some random seeds result in modal collapses and generate failure samples with many artifacts, which are excluded from the quantitative evaluation. This is also reported in a recent comprehensive study of GANs \cite{Lucic2017}.

\textbf{3-D vehicle shape completion:} We evaluate the results using the same metrics as in \cite{Stutz2018}, \ie Hamming distance. As we want to focus the evaluation on the fundamental differences between the different models and training strategies, here we mainly use the SN-clean dataset \cite{Stutz2018} (without noises) in the experiments of 3-D vehicle shape completion. The results are presented together in Table \ref{tab_main_compare}, with some samples visualized in Figure \ref{fig_shape_completion}. The quantitative results of the AML baseline method \cite{Stutz2018} in two conditions are reported (0.041 and 0.043) as well as the GAN baseline, which achieves the same performance (0.043). Our method achieves 0.035 Hamming distance and outperforms the baselines by a large relative margin. 

\subsection{Ablation studies}

We verify the effectiveness of our proposed training mechanism and show that three input-target pairs improve the performance. On both tasks, six different supervision cases are tested and the results are listed in Table \ref{tab_ablation}.

\textbf{2-D road layout hallucinating:} It can be observed that the performance degrades with different margins when ignoring different supervisions. In the case without the \textit{pre-selection pair} and \textit{masked prior knowledge pair}, the network works but with three metrics decreasing by a small margin. In the case of training with only the \textit{masked prior knowledge pair} or only the \textit{pre-selection pair}, the network is able to provide results with satisfactory segmentation accuracy, while together, the segmentation accuracy becomes optimal (78.8\% and 69.0\% mean IoU). This indicates that either pre-selection or masked prior knowledge pair can provide valid supervision on the unobserved regions independently, and together they can compensate for each other and provide optimal results in terms of segmentation accuracy. The performance without the \textit{observation pair} degrades in terms of the contour accuracy with certain margins (-2.4\% and -2.1\% $F$-measure), due to the fact that the observation pair provides a clear supervision at boundary regions. With three supervisions, the network exhibits the best contour accuracy and an optimal overall performance.

\textbf{3-D vehicle shape completion:} In terms of the 3-D vehicle shape completion task, similar conclusions can be drawn: with all the supervisions enabled, our approach achieves the optimal Hamming distance (0.035). Also, with only the \textit{masked prior knowledge pair} or the \textit{pre-selection pair} applied, the performance is already as good as the baselines (0.043) and even outperforms them (0.040). These two pairs are already sufficient for providing a valid supervision independently. However, we must clarify that there exists a major difference between the datasets of two tasks: unlike the 2-D road layout hallucinating, the prior knowledge and partial observation samples are from an identical distribution (ShapeNet dataset \cite{Chang2015}) in the 3-D vehicle shape completion task. This leads to some different observations in this ablation studies: 1) without the masked prior knowledge pair, the performance still remains optimal, and 2) adding the masked prior knowledge pair to the pre-selection pair degrades the performance. Due to the absence of the domain gap between the partially observed and prior knowledge dataset, the pre-selection pair provides a significantly stronger explicit supervision than the masked prior knowledge pair, which is not the case with more challenging data that have domain gaps, such as our 2-D road layout hallucinating benchmark.

\begin{table*}
	\renewcommand{\arraystretch}{1.2}
	\caption{Quantified performance of three approaches tested on the unseen KITTI dataset and the hold-out prior knowledge dataset.}
	\label{tab_generalization}
	\begin{center}
		\begin{tabular}{c|c||c|c|c}
			\hline
			{\multirow{2}{*}{\bfseries evaluated dataset}}	&{\multirow{2}{*}{\bfseries method}}		 & \multicolumn{1}{c|}{\bfseries contour }&\multicolumn{1}{c|}{\bfseries full pixels} &\multicolumn{1}{c}{\bfseries undetected pixels}\\
			& &  \bfseries $F$-measure&  \bfseries mean IoU & \bfseries mean IoU  \\
			\hline
			\hline
			\multirow{3}{*}{ \textbf{KITTI} (\cf Fig. \ref{fig_kitti_inputs})} &\bfseries AML & 21.5& 76.5 & 71.4  \\
			&\bfseries GAN & 39.2 &  83.4 & 76.0  \\
			&\bfseries ours &\bfseries 45.3& \bfseries 85.0 & \bfseries 78.5  \\
			\hline
			\hline
			\multirow{3}{*}{ \textbf{hold-out prior knowledge} (\cf Fig. \ref{fig_maksed_road_layout_inputs})} &\bfseries AML  & 14.2 & 70.1 & 64.3 \\
			&\bfseries GAN & 26.1 &  68.5 &  60.5 \\
			&\bfseries ours &\bfseries 39.9& \bfseries 85.1 & \bfseries 80.8 \\			
			\hline
		\end{tabular}
	\end{center}
\end{table*}

\subsection{Generalizability evaluation}

For 2-D road layout hallucinating task, we additionally evaluate the performance of three models on two unseen datasets, to verify the generalizability of different methods.

\begin{figure}[!tbp]
	\begin{center}
		\includegraphics[width=\linewidth]{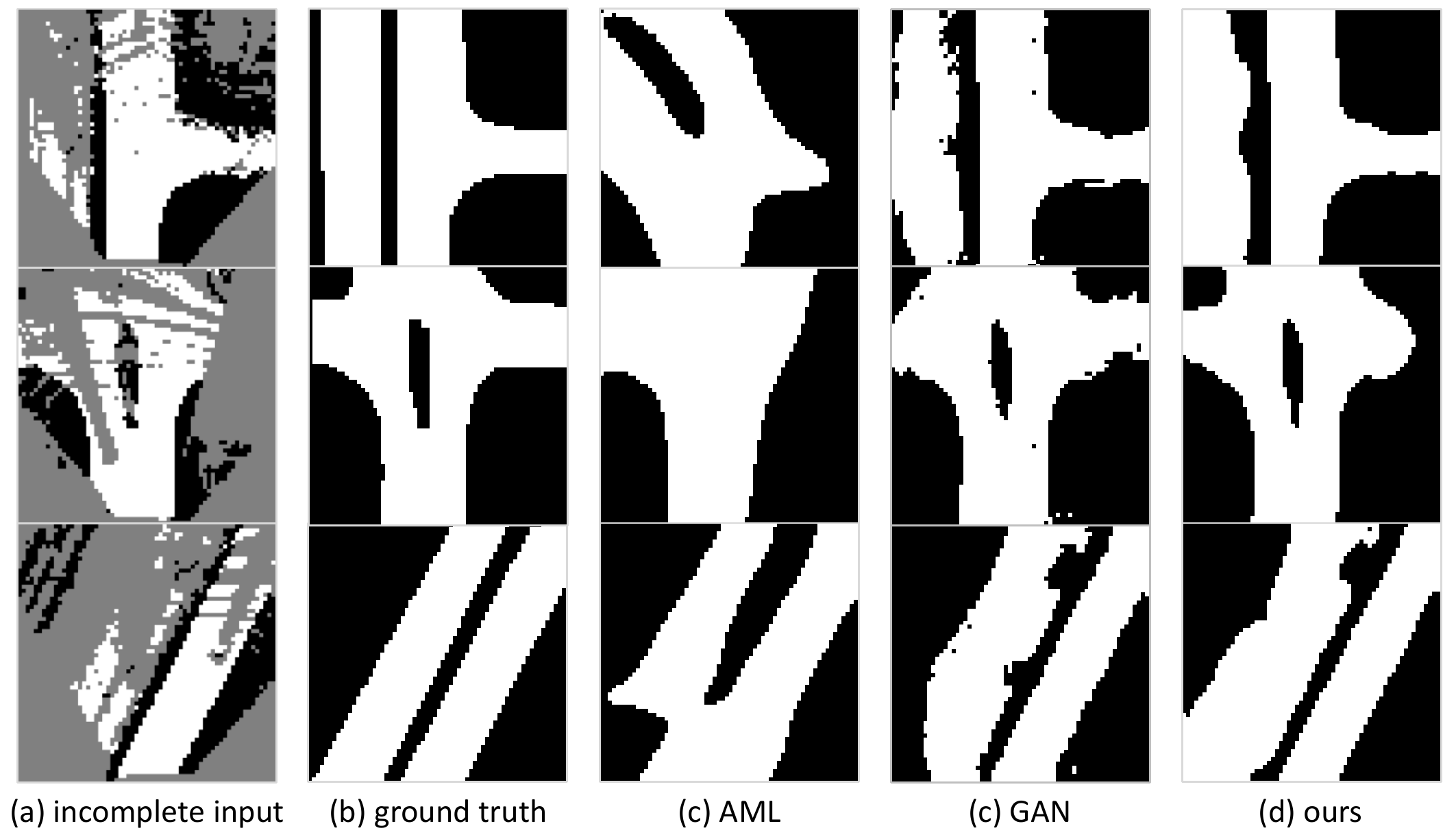}
	\end{center}
	\caption{Qualitative results on the unseen KITTI dataset.}
	\label{fig_kitti_inputs}
\end{figure}

\textbf{Unseen partially observed dataset:} The first generalizability experiment is conducted on the KITTI dataset which is never used in training. We feed the partially observed KITTI road layouts into the trained models and evaluate the quality of the predicted complete road layouts. As the pixel accuracy and IoU are highly correlated, we show the $F$-measure (contour accuracy) and mean IoU (segmentation accuracy) for simplicity. Figure \ref{fig_kitti_inputs} presents some visualized examples of the input incomplete maps and their corresponding predictions, and the upper part of Table \ref{tab_generalization} shows the quantified performance for three methods. We can observe from the KITTI input samples that 1) the camera's FOV is significantly larger than that in Cityscapes, and 2) the road region at large distance is severely missing, due to the sparsity of the provided depth map. These introduce domain gaps and causes the AML baseline method to fail in many examples. The qualitative and quantitative results show that the approaches with skip connections are more adaptive to the aforementioned domain gaps: they perform better compared to the AML baseline, in terms of more than 4.6\% mean IoU in both settings and 17.7\% $F$-measure. The main reason is that with skip connections, the shallow features are passed directly to the decoder, which makes the hallucinating robust against to unseen domain gaps. Moreover, with our proposed supervisions, the model exhibits noticeable performance improvements in all the metrics.

\begin{figure}[!tbp]
	\begin{center}
		\includegraphics[width=\linewidth]{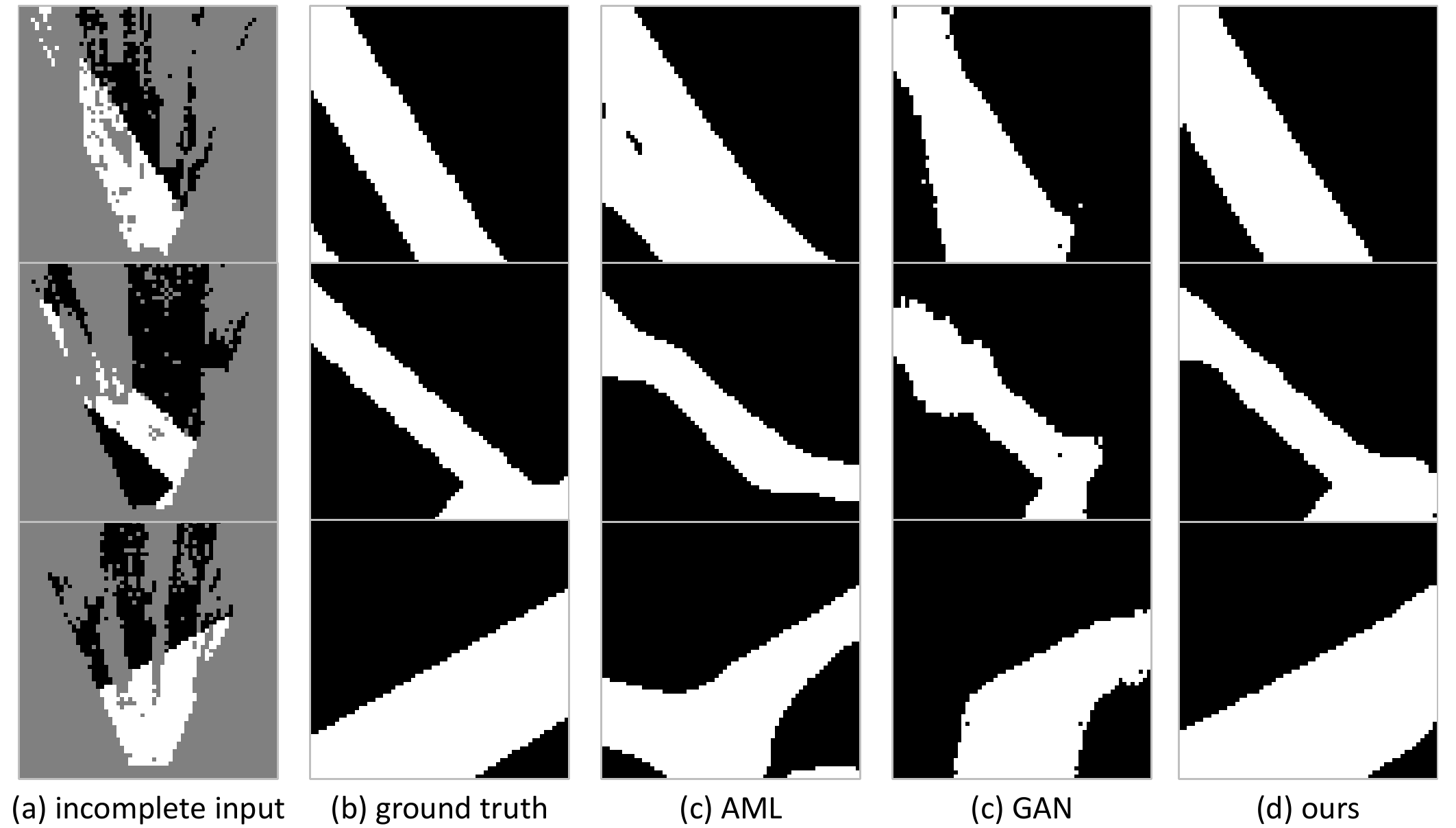}
	\end{center}
	\caption{Qualitative results on the hold-out masked prior knowledge set. The unobserved mask is from the Cityscapes test set.}
	\label{fig_maksed_road_layout_inputs}
\end{figure}

\textbf{Hold-out masked prior knowledge dataset:} To investigate this further, we generate another 1K prior knowledge dataset images and evaluated the performance of the baselines and our approach on these unseen road layouts. We manually apply masks sampled from the partially observed test set, and feed these incomplete maps into the learned models. The lower part of Table \ref{tab_generalization} shows the performance of the baselines and our approach, and Figure \ref{fig_maksed_road_layout_inputs} visualizes some hallucinating examples. Our approach exhibits the best performance, with a large margin compared to the other two methods (more than 13.8\% $F$-measure and  15\% mean IoU in both settings). The two datasets used in training do not follow an identical distribution, as they are from different domains: Cityscapes and AerialKITTI, and our network approach preserves the generalizability on the masked prior knowledge samples. Although this experiment has no relevance to the application of hallucinating road layouts, it does show that our training approach intrinsically generalizes better than the baselines.

\section{Conclusion}

In this work, we trained fully convolutional encoder-decoder networks with skip connections to successfully complete unobserved regions in road layout maps, as well as unobserved voxels in 3-D vehicle shapes. These two tasks can be seen as specific instances of hallucinating, when no ground truth is available for the partially observed domain, and the prior knowledge domain has no direct one-to-one correspondence to the partially observed domain. We demonstrate that our proposed single-step training strategy is key to unlocking the benefits of skip connections for this hallucination task. It outperforms two state-of-the-art baselines in multiple metrics, and it is significantly better in generalizing for unseen datasets. 

\bibliographystyle{IEEEtran}
\bibliography{./library}

\end{document}